# Predicting Blood Glucose with an LSTM and Bi-LSTM Based Deep Neural Network

Qingnan Sun, Marko V. Jankovic, Lia Bally, Stavroula G. Mougiakakou, *Member IEEE*

*Abstract* — A deep learning network was used to predict future blood glucose levels, as this can permit diabetes patients to take action before imminent hyperglycaemia and hypoglycaemia. A sequential model with one long-short-term memory (LSTM) layer, one bidirectional LSTM layer and several fully connected layers was used to predict blood glucose levels for different prediction horizons. The method was trained and tested on 26 datasets from 20 real patients. The proposed network outperforms the baseline methods in terms of all evaluation criteria.

## I. INTRODUCTION

Diabetes mellitus is a group of metabolic diseases characterised by hyperglycemia resulting from defects in insulin secretion, insulin action, or both [1]. Diabetes can be mainly classified into Type 1 diabetes (T1D) and Type 2 diabetes (T2D), where the former is caused by destruction of the pancreatic beta cells resulting in insulin deficiency, while the latter due to the ineffective use of insulin. The chronic hyperglycaemia of diabetes is associated with long-term damage, dysfunction, and failure of various organs, especially the eyes, kidneys, nerves, heart, and blood vessels [1]. Besides hyperglycaemia, hypoglycaemic events should also be prevented, as these can lead to unconsciousness, coma, and even death.

To keep blood glucose (BG) concentrations within a healthy range, T1D patients need external insulin, which can be delivered via syringes, insulin pens or insulin pumps [2]. This treatment can be improved by combining a pump for continuous subcutaneous insulin infusion (CSII) and a device for continuous glucose monitoring (CGM); this is known as sensor-augmented pump (SAP) therapy [3]. SAP therapy can be extended by introducing control algorithms to provide automatic insulin delivery. It is then referred to as the "Artificial Pancreas" (AP) system [4].

The CGM device can measure blood glucose level every 5 minutes, which means 288 measurements per day. This nearly continuous monitoring not only improves the evaluation of the current treatment, but also makes it possible to predict future blood glucose levels. If any hyperglycaemia or hypoglycaemia could be accurately predicted, the patients or the control algorithm could proactively take measures to avoid this.

Various methods have been investigated to predict blood glucose. Shanthi [5] introduced a classical statistical method, i.e. an autoregressive integrated moving average (ARIMA) model-based algorithm [6] to predict blood glucose in 30 to 60 minutes prediction horizons. Machine learning methods were introduced for blood glucose prediction, e.g. Daskalaki *et al.* [7], used a real-time learning recurrent neural network (RNN) fed with both glucose and insulin information. The model outperformed both an autoregressive (AR) model using glucose information, as well as an AR model with external insulin input (ARX); Bunescu *et al.* [8] proposed the use of support vector regression (SVR) for glucose prediction, by taking into consideration daily events, such as insulin boluses and meals; Georga *et al.* [9] proposed an approach which combined meal model, insulin model, exercise model and used SVR method to provide individualized glucose prediction. Deep learning methods recently proved to outperform the already established methodologies. This is due to the property of automatically extracting relevant features from the training samples in order to accurately predict the blood glucose. Mhaskar *et al.*[10] proposed a deep convolutional neural network (DCNN), which outperformed shallow network. One of the major challenges in designing systems using classical RNNs is their limited capacity to learn long-term dependencies, because of the vanishing or exploding gradient problem [11]. Recent deep RNNs incorporate mechanisms to address this problem [12], e.g. long-short-term memory (LSTM) which introduces the memory cell and forget gate into classical RNN network [13]. The memory cells enable the networks to improve prediction feasibility by combining its memories and the inputs, while the forget gate defines the information from the old state that can remain in the network. LSTM based networks have shown

Qingnan Sun is with the ARTORG Center for Biomedical Engineering Research, University of Bern. Murtenstrasse 50, CH-3008 Bern, Switzerland (phone: +41-31-632-7596; e mail: qingnan.sun@artorg.unibe.ch).

Marko V. Jankovic is with the Department of Emergency Medicine, Bern University Hospital "Inselspital", Bern, Switzerland and with the ARTORG Center for Biomedical Engineering Research, University of Bern. Murtenstrasse 50, CH-3008 Bern, Switzerland (phone: +41 31 632 75 96; fax: +41 31 632 75 76; e-mail: marko.jankovic@artorg.unibe.ch).

Lia Bally is with the Department of Diabetes, Endocrinology, Clinical Nutrition and Metabolism, Inselspital, Bern University Hospital, University of Bern, Bern, Switzerland and with the Department of General Internal Medicine, Inselspital, Bern University Hospital, University of Bern. Freiburgstrasse 15, CH-3010 Bern, Switzerland (phone: +41-31-632-4070; e mail: lia.bally@insel.ch).

Stavroula G. Mougiakakou is with the ARTORG Center for Biomedical Engineering Research, University of Bern, Murtenstrasse 50, CH-3008 Bern, Switzerland and the Department of Diabetes, Endocrinology, Clinical Nutrition and Metabolism, Bern University Hospital "Inselspital", Bern, Switzerland (phone: +41-31-632-7592; e mail: stavroula.mougiakakou@artorg.unibe.ch).

promising results for time series prediction, and have been applied to predict stock prices [14], highway trajectories [15], sea surface temperatures [16], or to learn the physiological models of blood glucose behaviour [17] etc. LSTM can learn much faster than other networks and solve complex tasks that have never been solved by previous recurrent network algorithms [13]. By incorporating with a bidirectional structure, each cell of LSTM is enabled to access the context from the both past and future directions. A deep bidirectional LSTM (Bi-LSTM) structure was introduced for blood pressure prediction by Su *et al.* [18].

In this work, we focus on using a LSTM and bidirectional LSTM based deep neural network to predict blood glucose levels based on CGM measurements. As baseline, we used the prediction results from the ARIMA model and SVR method. In this preliminary study, only blood glucose data were considered as feature for the prediction.

## II. METHODOLOGY

In this section, the methods and models used are briefly presented. The setup of the prediction model and training phase are also introduced.

### A. LSTM network

LSTM is a variant of an RNN network, as proposed by Hochreiter and Schmidhuber in 1997 [13]. The LSTM network solves the long-term dependency problem that occurs in classical RNNs, by introducing memory *C* and the gate structure.

Fig.1. illustrates the structure of the LSTM cell [19], which has four gates, i.e. input gate *i*, forget gate *f*, control gate *c*, and output gate *o*.

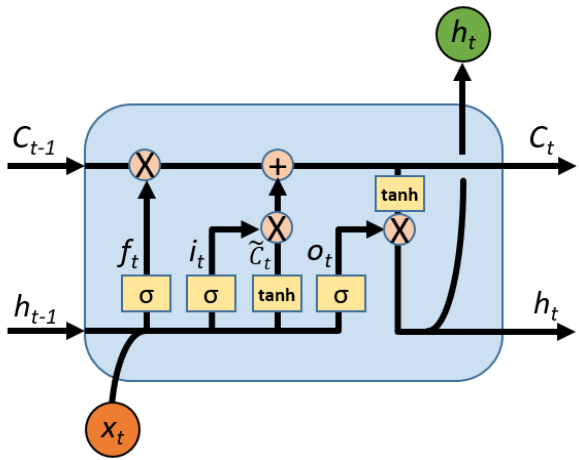

Fig. 1. Structure of the LSTM cell

The input gate decides which information can be transferred to the cell, which can be defined as:

$$i_t = \sigma(W_i \cdot [h_{t-1}, x_t] + b_i) \quad (1)$$

The forget gate decides which information from input should be neglected from the previous memory and is defined as:

$$f_t = \sigma(W_f \cdot [h_{t-1}, x_t] + b_f) \quad (2)$$

The control gate controls the update of cell state from $C_{t-1}$ to $C_t$, based on equations (3) and (4).

$$\tilde{C}_t = tanh(W_C \cdot [h_{t-1}, x_t] + b_C) \quad (3)$$
$$C_t = f_t * C_{t-1} + i_t * \tilde{C}_t \quad (4)$$

The output gate is responsible for generating the output and updating the hidden vector $h_{t-1}$. This process can be defined as:

$$o_t = \sigma(W_o \cdot [h_{t-1}, x_t] + b_o) \quad (5)$$
$$h_t = o_t * \tanh(C_t) \quad (6)$$

In equations (1) to (6) $\sigma$ is the sigmoid activation function, the *W*s are the corresponding weight matrices, and tanh is used to scale the values into the range -1 to 1.

### B. Bidirectional LSTM network

Bidirectional recurrent neural networks (BRNN) were first introduced by Schuster and Paliwal in 1997. BRNN can be trained using all available input information in the past and future of a specific time [20]. By splitting the state neurons of a regular RNN into the positive and negative time directions, the network isolates the outputs from forward states and backward states. During the training process, BRNN is trained in both forward and backward directions. Fig.2. shows the general structure of BRNN. Bidirectional structure can be applied to the variants of the RNNs; in this work we used bidirectional LSTM.

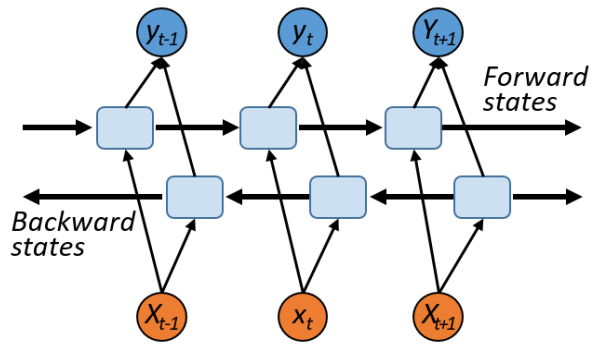

Fig. 2. General structure of BRNN

### C. Prediction model and training setup

We built the prediction model with the high-level neural networks API Keras version 2.0.8 in the Python 3.4.3 environment.

We considered one LSTM layer and one bidirectional LSTM layer, each with 4 units, and three fully connected layers with 8, 64 and 8 units, respectively. The output layer, the one unit dense layer, was used to output the final predicted blood glucose value. Fig.3. shows the input and output dimension of each layer. The number of train epochs was fixed by 100 iterations.

Two rounds of pre-train, with *in silico* data as well as real patient data, were introduced to generate a generalised "global model". For the pre-train phase, the epoch number was determined by running the experiments with 100 to 2600 iterations (increment is 100 iterations), and the model that combines good performance and low numbers of iteration is chosen. This will be described in detail in section IV. Cross validation (data split: 67%, 33%) was used to prevent over-fitting during pre-train and train phases.

## III. DATASETS AND EVALUATION CRITERIA

Both real patients datasets and *in silico* datasets were used in this work.

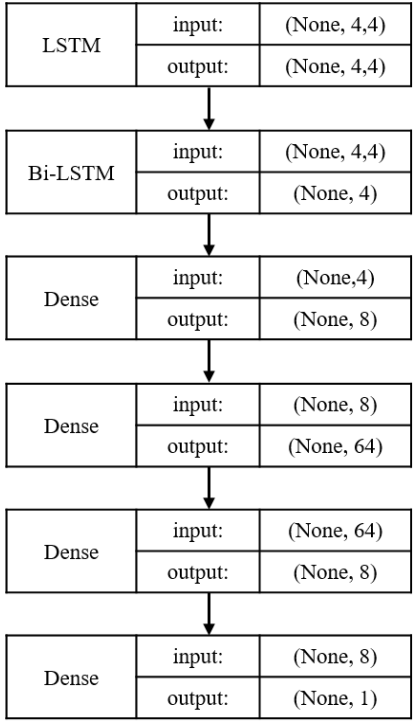

Fig. 3. Input and output dimensions of each layer

*A. Datasets from real patients*

The datasets, which were described in [21], were from a pilot prospective randomised controlled crossover study (NCT02546063) of GoCARB [22]. In total, 20 adults with T1D (mean age 35 ± 14 years, duration of diabetes 17 ± 10 years) were involved. In this work, we only used the CGM measurements, which were measured with 5 minutes sampling time. The 20 datasets contain 6975 ± 1612 measurements.

The datasets were pre-processed to overcome i) single outliers between two normal measurements, ii) lack of measurements for a period. Linear interpolation was used to calculate the missing data in situation i). As in future work more features will be involved in prediction, the original datasets for situation ii) were divided into sub-datasets, which have valid measurements of all the variables (i.e. CGM measurements, basal rate, bolus amount etc.).

Then we chose the sub-datasets with the minimum length of 1500 measurements (at least 5 days). On the basis of this pre-processing and selection, in total 26 sub-datasets (length: 1791 ± 141) could meet the requirements. The remaining sub-datasets, which have fewer than 1500 measurements, were merged into a single data file that was used to pre-train the network.

*B. Datasets from in silico trials*

We used the FDA-accepted UVa/Padova T1D Simulator to establish 38-day *in silico* clinical trials for 11 virtual adult subjects. A virtual CGM with sampling time 5 minutes was used; the glucose measurements were extracted and merged into a single file for pre-train purpose. TABLE I shows the overview of the datasets.

TABLE I: OVERVIEW OF DATASETS

|  | Simulation data | Real data 1 (length<1500) | Real data 2 (length>1500) |
| --- | --- | --- | --- |
| Usage | 1st round pre-train | 2nd round pre-train | 67% - train 33% - test |
| Number of Datasets | 1 | 1 | 26 |
| Samples | 120395 | 93443 | 1791±141 |
| Mean Glucose (mg/dL) | 131.29 | 161.45 | 154.9±25.7 |
| Hypo-Number | 33 | 234 | 9±8 |
| Hyper-Number | 616 | 640 | 19±7 |

*Hypo: Glucose value lower than 70 mg/dl*
*Hyper: Glucose value above 180 mg/dl*

*C. Evaluation criteria*

Root mean square error (RMSE, [mg/dl]), correlation coefficient (CC), time lag (TL, [min]) and fit were used to evaluate the prediction performance. The evaluation criteria indicate the overall prediction ability of the models by comparing the test datasets and the predicted value.

RMSE indicates the difference between the target data and the predicted data. This is calculated by taking the square root of the mean of the square of all the errors (7). A lower RMSE means better prediction performance.

$$\text{RMSE} = \sqrt{E((G-\hat{G})^2)} = \sqrt{\frac{1}{N}\sum(G-\hat{G})^2} \quad (7)$$

where $G$ and $\hat{G}$ are the actual and predicted glucose values respectively.

CC represents the linear dependence between two datasets and is calculated by

$$CC_{xy} = \frac{\sigma_{xy}}{\sigma_x \sigma_y} \quad (8)$$

where $\sigma_x$ and $\sigma_y$ are the standard deviations and $\sigma_{xy}$ is the covariance. For our work, CC could be further described as:

$$CC = \frac{\sum(G-G_{mean})(\hat{G}-\hat{G}_{mean})}{\sqrt{\sum(G-G_{mean})^2}\sqrt{\sum(\hat{G}-\hat{G}_{mean})^2}} \quad (9)$$

where $G_{mean}$ and $\hat{G}_{mean}$ are the mean value of $G$ and $\hat{G}$, respectively.

TL is defined as the minimum time-shift between the actual and predicted signal which gives the highest correlation coefficient between them [23].

Fit is calculated on the basis of the fraction of RMSE and root mean square difference between target and its mean value. Equation (10) shows the calculation in detail. For a time series, a higher Fit value indicates better prediction performance of the algorithm.

$$Fit = \left(1 - \frac{\sqrt{\frac{1}{N}\sum(G-\hat{G})^2}}{\sqrt{\frac{1}{N}\sum(G-G_{mean})^2}}\right) * 100\% \quad (10)$$

## IV. EXPERIMENTAL RESULTS

ARIMA, SVR and LSTM methods were used to predict the upcoming blood glucose levels for prediction horizons (PH) of 15, 30, 45 and 60 minutes. TABLE II presents the prediction results, and the best performance are shown in bold, i.e. the lowest RMSE and Time Lag, and the largest CC and Fit. The LSTM method outperforms the classic methods in all the PHs with respect to all the evaluation criteria. It is promising that, in comparison with ARIMA and SVR, the LSTM method can simultaneously decrease RMSE and Time Lag, while CC and Fit are both increased.

TABLE II: Prediction results of different methods

| PH = 15 minutes | | | | |
|---|---|---|---|---|
| *Methods* | *RMSE* | *CC* | *Time Lag* | *Fit* |
| ARIMA | 12.256 | 0.972 | 10.192 | 76.425 |
| SVR | 11.694 | 0.973 | 9.808 | 77.565 |
| **LSTM** | **11.633** | **0.974** | **9.423** | **77.714** |
| PH = 30 minutes | | | | |
| *Methods* | *RMSE* | *CC* | *Time Lag* | *Fit* |
| ARIMA | 22.924 | 0.903 | 22.885 | 55.923 |
| SVR | 22.135 | 0.904 | 20.769 | 57.644 |
| **LSTM** | **21.747** | **0.909** | **20.385** | **58.523** |
| PH = 45 minutes | | | | |
| *Methods* | *RMSE* | *CC* | *Time Lag* | *Fit* |
| ARIMA | 32.588 | 0.806 | 37.885 | 37.463 |
| SVR | 30.628 | 0.812 | 34.423 | 41.595 |
| **LSTM** | **30.215** | **0.818** | **32.692** | **42.563** |
| PH = 60 minutes | | | | |
| *Methods* | *RMSE* | *CC* | *Time Lag* | *Fit* |
| ARIMA | 40.841 | 0.698 | 52.885 | 21.694 |
| SVR | 37.422 | 0.709 | 47.885 | 28.893 |
| **LSTM** | **36.918** | **0.722** | **46.346** | **30.079** |

For the LSTM method, the epoch number for the two rounds of pre-train process was determined by running the experiment with epoch number from 100 to 2000, with increment of 100. TABLE III shows the results of the experiments for PH=30 minutes as an example. When performance deteriorates, the background colour of the value changes from dark green to dark red. With some epoch numbers, e.g. 900, 1300, 1500, 1700, etc., the performance is better with respect to all the indices. By taking into consideration the balance between performance and time cost for pre-train, 1300 was chosen as the pre-train epoch number for PH=30 minutes. The pre-train epoch numbers for PH=15, 45 and 60 are 900, 1200 and 1800.

## V. CONCLUSION

In the present study, an LSTM network with one LSTM layer, one bi-directional LSTM layer along with several fully connected layers were used to predict blood glucose concentration. The LSTM network was pre-trained with both *in silico* data and real patient data to generate a "global model". Then the model was trained and tested with 26 real datasets, with each dataset containing more than 5-day CGM data. The epoch number for the training phase was low, which ensures that the method could work rapidly, even on a mobile platform. On the basis of the evaluation criteria, the LSTM network outperformed the baseline methods ARIMA and SVR. In comparison with the baselines and in all prediction horizons, the LSTM network reduced the RMSE and TL, while CC and Fit were increased.

TABLE III: Prediction performance of LSTM method with different pre-train epochs (for PH=30 minutes)

| *Epochs* | *RMSE* | *CC* | *Time Lag* | *Fit* |
|---|---|---|---|---|
| **100** | 22.649 | 0.902 | 21.731 | 56.749 |
| **200** | 22.509 | 0.904 | 21.539 | 57.077 |
| **300** | 22.317 | 0.904 | 21.346 | 57.448 |
| **400** | 22.140 | 0.906 | 20.577 | 57.778 |
| **500** | 22.021 | 0.907 | 20.962 | 58.004 |
| **600** | 22.014 | 0.907 | 20.577 | 58.010 |
| **700** | 22.064 | 0.906 | 20.769 | 57.918 |
| **800** | 22.035 | 0.907 | 20.962 | 58.039 |
| **900** | 21.828 | 0.908 | 20.577 | 58.338 |
| **1000** | 22.113 | 0.905 | 20.769 | 57.760 |
| **1100** | 22.137 | 0.907 | 20.962 | 57.813 |
| **1200** | 22.082 | 0.907 | 20.962 | 57.933 |
| **1300** | 21.747 | 0.909 | 20.385 | 58.523 |
| **1400** | 22.025 | 0.906 | 20.962 | 57.967 |
| **1500** | 21.931 | 0.908 | 20.577 | 58.173 |
| **1600** | 22.008 | 0.907 | 20.577 | 58.070 |
| **1700** | 21.804 | 0.908 | 20.385 | 58.429 |
| **1800** | 21.850 | 0.908 | 20.577 | 58.343 |
| **1900** | 21.848 | 0.907 | 20.769 | 58.315 |
| **2000** | 21.982 | 0.907 | 20.769 | 58.113 |

Since only CGM measurements were needed, the method could be used for patients using oral drugs, insulin pens or the CSII pump. In the next step, we will include more features to improve the performance of the method for specific target user groups. An alarm mechanism will be implemented to detect the up-coming hyper- and hypoglycaemic events, on the basis of the predicted glucose concentrations.

REFERENCES

[1] A. D. Association, "Diagnosis and classification of diabetes mellitus," *Diabetes Care*, vol. 32, no. SUPPL. 1, 2009.
[2] JDRF, "Type 1 Diabetes Treatments." [Online]. Available: http://www.jdrf.org/about/what-is-t1d/treatment.
[3] A. Bergenstal, Richard M, Tamborlane, William v, ahmann, "Sensor-Augmented Pump Therapy for A1C Reduction (STAR 3) Study," *Diabetes Care*, vol. 34, no. Star 3, pp. 2403–2405, 2011.
[4] D. Bruttomesso *et al.*, "Closed-Loop Artificial Pancreas Using Subcutaneous Glucose Sensing and Insulin Delivery and a Model Predictive Control Algorithm: Preliminary Studies in Padova and Montpellier," *J. Diabetes Sci. Technol.*, vol. 3, no. 5, pp. 1014–1021, 2009.
[5] S. Shanthi, "Advanced Engineering Informatics A novel approach for the prediction of glucose concentration in type 1 diabetes ahead in time through ARIMA and differential evolution," vol. 38, pp. 4182–4186, 2011.
[6] J. Yang, L. Li, Y. Shi, and X. Xie, "An ARIMA Model with Adaptive Orders for Predicting Blood Glucose Concentrations and Hypoglycemia," vol. 2194, no. 1, 2018.

[7] E. Daskalaki, A. Prountzou, P. Diem, and S. G. Mougiakakou, "Real-Time Adaptive Models for the Personalized Prediction of Glycemic Profile in Type 1 Diabetes Patients," *Diabetes Technol. Ther.*, vol. 14, no. 2, pp. 168–174, 2012.

[8] R. Bunescu, N. Struble, C. Marling, J. Shubrook, and F. Schwartz, "Blood Glucose Level Prediction Using Physiological Models and Support Vector Regression," *2013 12th Int. Conf. Mach. Learn. Appl.*, pp. 135–140, 2013.

[9] E. I. Georga, V. C. Protopappas, and D. I. Fotiadis, "Glucose Prediction in Type 1 and Type 2 Diabetic Patients Using Data Driven Techniques," *Knowledge-Oriented Appl. Data Min.*, pp. 277–296, 2011.

[10] H. N. Mhaskar, S. V. Pereverzyev, and M. D. van der Walt, "A Deep Learning Approach to Diabetic Blood Glucose Prediction," *Front. Appl. Math. Stat.*, vol. 3, no. July, pp. 1–11, 2017.

[11] Y. Bengio, P. Simard, and P. Frasconi, "Learning Long-Term Dependencies with Graident Descent is Difficult," *Saudi Med J*, vol. 33, pp. 3–8, 2012.

[12] S. Park, S. Min, H.-S. Choi, and S. Yoon, "Deep Recurrent Neural Network-Based Identification of Precursor microRNAs," *Nips*, no. Nips, 2017.

[13] S. Hochreiter and J. Schmidhuber, "Long Short-Term Memory," *Neural Comput.*, vol. 9, no. 8, pp. 1735–1780, 1997.

[14] M. Roondiwala, H. Patel, and S. Varma, "Predicting Stock Prices Using LSTM," vol. 6, no. 4, pp. 2015–2017, 2017.

[15] F. Altché, A. De, and L. Fortelle, "An LSTM Network for Highway Trajectory Prediction."

[16] Q. Zhang, H. Wang, J. Dong, G. Zhong, and X. Sun, "Prediction of Sea Surface Temperature using Long Short-Term Memory," pp. 1–5, 2017.

[17] S. Mirshekarian, R. Bunescu, C. Marling, and F. Schwartz, "Using LSTMs to learn physiological models of blood glucose behavior," *Proc. Annu. Int. Conf. IEEE Eng. Med. Biol. Soc. EMBS*, pp. 2887–2891, 2017.

[18] P. Su, X. Ding, Y. Zhang, F. Miao, and N. Zhao, "Learning to Predict Blood Pressure with Deep Bidirectional LSTM Network," pp. 1–19, 2017.

[19] C. Olah, "Understanding LSTM Networks," 2015. [Online]. Available: http://colah.github.io/posts/2015-08-Understanding-LSTMs/.

[20] M. Schuster and K. K. Paliwal, "Bidirectional recurrent neural networks," *IEEE Trans. Signal Process.*, vol. 45, no. 11, pp. 2673–2681, 1997.

[21] B. Lia *et al.*, "Carbohydrate Estimation Supported by the GoCARB system in Individuals With Type 1 Diabetes: A Randomized Prospective Pilot Study," *Diabetes Care*, vol. 40, no. 2, pp. e6–e7, 2017.

[22] M. Anthimopoulos *et al.*, "Computer vision-based carbohydrate estimation for type 1 patients with diabetes using smartphones," *J. Diabetes Sci. Technol.*, vol. 9, no. 3, pp. 507–515, 2015.

[23] E. Daskalaki, "Towards the External Artificial Pancreas : Design and Development of a Personalized Control System for Glucose Regulation in Individuals with Type 1 Diabetes," University of Bern, 2013.